\documentclass[10pt,twocolumn,letterpaper]{article}

\usepackage{cvpr}              %

\usepackage{graphicx}
\usepackage{amsmath}
\usepackage{amssymb}
\usepackage{booktabs}
\usepackage{xcolor}
\usepackage{color, colortbl}
\usepackage[accsupp]{axessibility}  %

\usepackage{multirow}
\usepackage{array}
\newcommand{\PreserveBackslash}[1]{\let\temp=\\#1\let\\=\temp}
\newcolumntype{C}[1]{>{\PreserveBackslash\centering}p{#1}}
\newcolumntype{L}[1]{>{\PreserveBackslash\raggedright}p{#1}}
\usepackage{amssymb}%
\usepackage{pifont}%
\newcommand{\cmark}{\ding{51}}%
\newcommand{\xmark}{\ding{55}}%

\newcommand\blfootnote[1]{%
  \begingroup
  \renewcommand\thefootnote{}\footnote{#1}%
  \addtocounter{footnote}{-1}%
  \endgroup
}

 \newcommand{\ccCR}[1]{{\color{black}{}#1}}

\newcommand{\KAcamera}[1]{{\color{black}{}#1}}

\newcommand{\modelname}[0]{{HierVL}}
\definecolor{Gray}{gray}{0.9}

\usepackage[pagebackref,breaklinks,colorlinks]{hyperref}

\usepackage[capitalize]{cleveref}
\crefname{section}{Sec.}{Secs.}
\Crefname{section}{Section}{Sections}
\Crefname{table}{Table}{Tables}
\crefname{table}{Tab.}{Tabs.}

\def\cvprPaperID{8470} %
\def\confName{CVPR}
\def\confYear{2023}

\begin{document}

\title{\modelname: Learning Hierarchical Video-Language Embeddings}

\author{Kumar Ashutosh$^{1}$, Rohit Girdhar$^{2}$, Lorenzo Torresani$^{2}$, Kristen Grauman$^{1,2}$\\
$^{1}$UT Austin, $^{2}$FAIR, Meta AI\\
}
\maketitle

\begin{abstract}
\blfootnote{Website: \href{https://vision.cs.utexas.edu/projects/hiervl/}{https://vision.cs.utexas.edu/projects/hiervl/}}
Video-language embeddings are a promising avenue for injecting semantics into visual representations, but existing methods capture only short-term associations between seconds-long video clips and their accompanying text. 
We propose \modelname, a novel hierarchical video-language embedding 
that simultaneously accounts for both long-term and short-term associations. 
As training data, we take videos accompanied by timestamped text descriptions of human actions, together with a high-level text summary of the activity throughout the long video (as are available in Ego4D).
We introduce a hierarchical contrastive training objective that encourages text-visual alignment at both the clip level and video level.
While the clip-level constraints use the step-by-step descriptions to capture  \emph{what} is happening in that instant, the video-level constraints use the summary text to capture \emph{why} it is happening, i.e., the broader context for the activity and the intent of the actor.
Our hierarchical scheme
yields a clip representation that outperforms its single-level counterpart as well as a long-term video representation that achieves SotA results on tasks requiring long-term video modeling.  HierVL
successfully transfers to multiple challenging downstream tasks (in EPIC-KITCHENS-100, Charades-Ego, HowTo100M) in both zero-shot and fine-tuned settings.

\end{abstract}
\section{Introduction}
\label{sec:intro}

Understanding human activity in video is a fundamental vision problem with abundant applications in augmented reality, robotics, and information retrieval.  The field has made exciting advances, from new models for recognition~\cite{mvitv2,memvit,slowfast} and self-supervised representations~\cite{mil-nce,univl,videoclip,egovlp}  to major datasets~\cite{ego4d,howto100m,epic-kitchens-100,crosstask,charades-ego}.  Nonetheless, activity understanding in video lags noticeably behind object understanding in images, where today's 
AI models compete well with people.

\begin{figure}[t]
\centering
\includegraphics[width=0.47\textwidth]{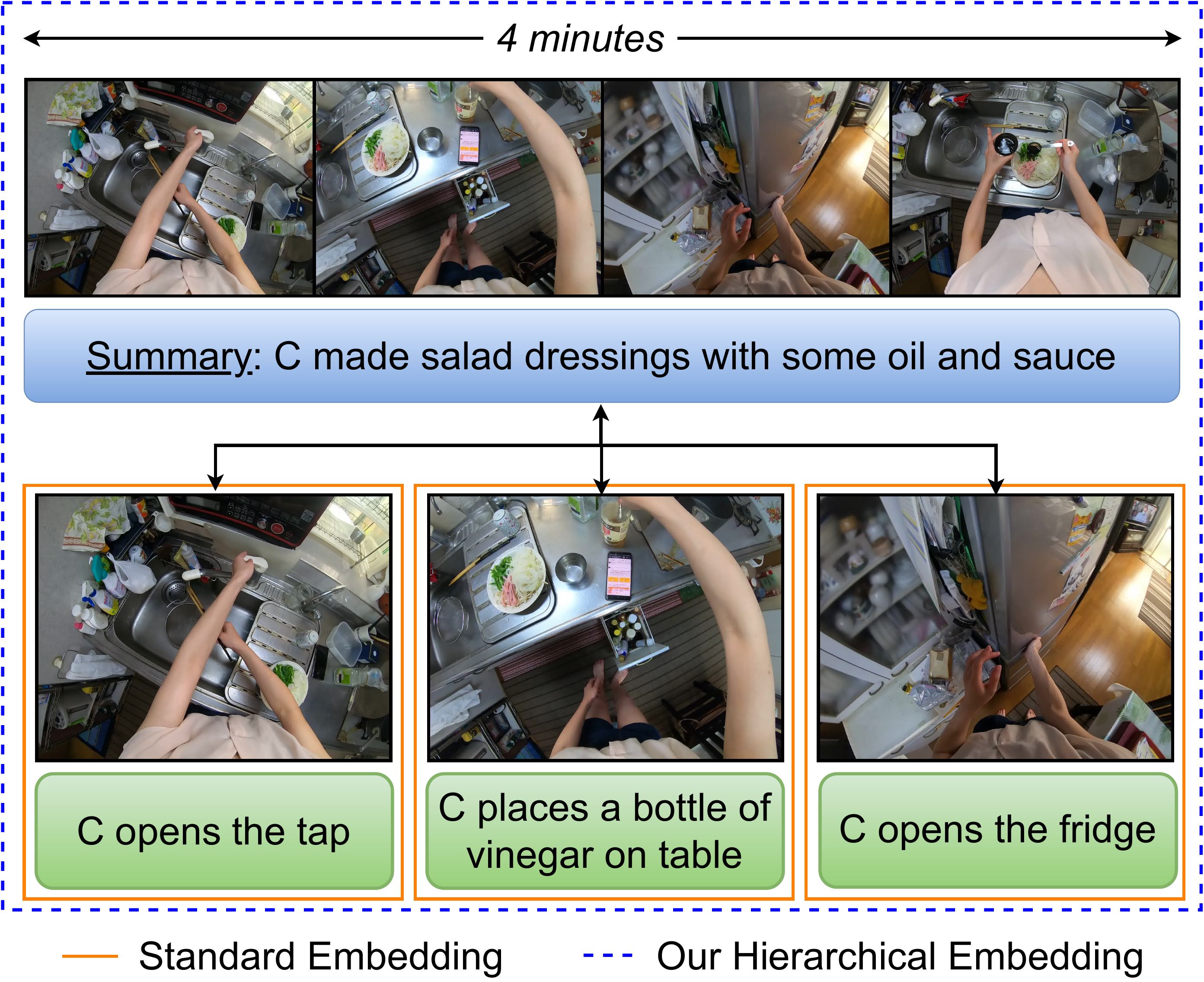}
\caption{Conventional video-language embeddings are trained to match short-term clips with their corresponding descriptions, e.g., open tap (in orange boxes), thus capturing \emph{what is happening}. Our hierarchical video-language embedding (in dotted blue box) learns both short-term and long-term visual-text relations, thereby capturing \emph{why is it happening} (e.g., making salad dressing). Long-term intent is conveyed by textual summaries (blue) that give an abstractive summary of the whole video, and complement the more literal step-by-step narrations (green).
}%
\label{fig:motivation}
\vspace{-0.10in}
\end{figure}

One key reason for this discrepancy is the fact that whereas objects present themselves directly %
in the pixels---no subtext required---activity naturally has broad temporal context rooted in the human actor's (latent) intentions.  Not only does an activity stretch across video frames, but also its interpretation  relies on the larger context of what the person is trying to accomplish.  Thus, there is a natural \emph{hierarchy} of information in video, starting with the short-term ``what the person is literally doing right now" (e.g., reaching for the stove) and going all the way to the long-term ``what the person aims to do" (e.g., cook dinner).

As a step towards capturing this hierarchy, we explore video-language representation learning. %
Video often has accompanying timestamped text, whether from spoken narrations in a how-to video~\cite{howto100m,coin,crosstask}, closed caption text and scripts~\cite{activitynet,movieqa}, or  deliberate text annotations~\cite{msr-vtt,epic-kitchens-100,ego4d}.  
Existing video-language models learn a correspondence between the two modalities by matching short video segments with their text counterpart, typically with a learned embedding~\cite{egovlp,mil-nce,videoclip,frozenintime} that produces a language-enriched video clip encoder.  
However, this standard approach risks capturing only the short-term actions. Granular comments such as \textit{``now I pour milk in the pan"} or \textit{``he picked up a water hose"} fail to capture the overall goal of the activity, like \textit{making a coffee} or \textit{cleaning a car}.  As a result, at inference time their encodings for unseen videos can be myopic and miss sequential dependencies between observed events.

To tackle this problem, we introduce \modelname: a novel hierarchical video-language model that captures both short-term actions and long-term intents in video.  Unlike standard video-language embeddings, our method %
aims to simultaneously capture the immediate observed actions as well as their contribution to the longer-term goal.
To that end, given training video accompanied by timestamped clip-level text descriptions as well as global (video-level) text \emph{summaries}, \modelname~learns a video-text embedding for hierarchical temporal understanding using two layers of contrastive learning.  The top (parent) layer encourages the \emph{aggregated video clips} to be close to the overarching textual summary (e.g., \emph{he makes spaghetti dinner}), while the bottom (child) layer trains individual clips to be similar to their respective descriptions (e.g., \emph{he turns on the cooker}).  
See~\cref{fig:motivation}.

To our knowledge, ours is the first work to create a hierarchical video-language embedding.  Our idea to blend abstract textual summaries with literal text descriptions is new.  Furthermore, our model design addresses constituent technical challenges---namely, we circumvent the typical expense of long-term feature learning~\cite{memvit,scsampler,clip-hitchhiker} by using aggregation of short-term features, and we show how to jointly train with two levels of annotation in a way that staves off catastrophic forgetting of either layer.

This hierarchical training yields not only global video-level representations that capture long-term information (e.g., intent and temporal dependencies), but also clip-level video features that are more expressive than those traditionally learned via single-level schemes. This happens by means of our parent-child learning framework, which requires the aggregation of clip features within a video to match the long-term context captured by the summary. 

We demonstrate our model by training with the narrations and summaries in the 3,670-hour egocentric video dataset Ego4D~\cite{ego4d,ego4dcons}.  
We show that \modelname~outperforms strong baselines and state-of-the-art methods for multiple video benchmarks, successfully transferring its pretrained representation for inference on Charades-Ego \cite{charades-ego}, EPIC-KITCHENS \cite{epic-kitchens-100}, and HowTo100M \cite{howto100m}.%
\footnote{Note that we do not need any text or summary annotations for these downstream datasets and tasks.}
We evaluate our representations on both hierarchy levels.
In particular, at the time of submission, \modelname~achieves state-of-the-art performance on Ego4D Long Term Anticipation (LTA), Charades-Ego Action Recognition, EPIC-KITCHENS-100 Multi-Instance Retrieval (zero-shot and fine-tuned settings), and HowTo100M Long Video Classification.

\section{Related Work}
\label{sec:related}

\textbf{Activity recognition and detection. %
} %
Video understanding spans tasks like action recognition
\cite{omnivore,mvitv2,uniformer,memvit,slowfast}, action anticipation \cite{avt,rulstm,intention,whenwillyoudowhat,gao2017red}, procedure learning \cite{chang2020procedure,bansal2022my,naing2020procedure,bi2021procedure,zhao2022p3iv}, and action localization \cite{videoclip,taco,crosstask,actionformer}.  
Various video datasets facilitate research in these directions, including Internet video collections like HowTo100M \cite{howto100m}, YouCookII \cite{youcook2}, and CrossTask \cite{crosstask}, %
as well as freshly recorded 
 datasets like CharadesEgo~\cite{charades-ego}, EPIC-KITCHENS~\cite{epic-kitchens-100}, and Ego4D~\cite{ego4d,ego4dcons}. 
As a training resource, we use Ego4D~\cite{ego4d,ego4dcons}, a large-scale diverse collection of in-the-wild wearable camera videos of daily-life activity around the world.  The Ego4D videos have low-level text descriptions (``narrations") of every action performed by the camera wearer, as well as video-level summaries, making them well-suited for our idea.

\textbf{Long-form video representations.} %
Longer videos introduce computational bottlenecks, making long-form video understanding challenging. %
There are several workarounds to make the task computationally feasible. Traditional methods include using pre-computed features that minimize backpropagation requirements \cite{pre-computed-1,pre-computed-2,pre-computed-3,pre-computed-4,long-form-video-understanding} or \KAcamera{decreasing the frame-rate %
\cite{more-frame-1,more-frames-2,more-frames-3,scsampler,more-frames-4,more-frames-5,merlot,clipbert,violet}}. Recent methods mitigate the computational requirements by creating a ``feature-bank" \cite{ltfb} or caching memory~\cite{memvit}.  Structured state space sequence models (S4) \cite{gu-s4, gedas-eccv22} reduce the quadratic complexity of self-attention to linear, enabling efficient training of long-sequence tasks. Another promising approach is to aggregate fine-grained clip-level features \cite{clip-hitchhiker,aggregation-1,aggregation-2,aggregation-3,aggregation-4,aggregation-5,aggregation-6,aggregation-7,Wang_2022_CVPR} into an overall video representation, as typically employed for video classification tasks. While all these methods are video-only, we propose a multi-modal long-form representation for both visual and textual modalities.

\textbf{Joint video and language learning.} The idea of projecting visual and language representations in the same embedding space is widely used for multi-modal understanding \cite{howto100m,mil-nce,frozenintime,egovlp,videoclip}.  Such joint representations enable several tasks, like language grounding in images \cite{img-grounding-1,img-grounding-2,img-grounding-3,img-grounding-4,img-grounding-5}, %
image captioning \cite{captioning-1,captioning-2,captioning-3,captioning-4,captioning-5}, and image retrieval \cite{img-retrieval-1,img-retrieval-2,img-retrieval-3,img-retrieval-4,img-retrieval-5}, as well as %
text-to-video retrieval \cite{vid-retrieval-1,vid-retrieval-2,vid-retrieval-3,videoclip,univl}, video captioning \cite{univl,coot,xu2021vlm,omnivl}, and video question answering \cite{vqa-1,hero,tvqa+,just-ask,vqa-2,naq}. %
Several of these methods \cite{howto100m,mil-nce,videoclip,egovlp,univl} use contrastive learning  (e.g., InfoNCE \cite{infonce}) and match video clips (or images) with their narrations (or captions) in a self-supervised manner. %
The self-supervised model in~\cite{brave} uses both narrow and broad windows of visual and audio, and focuses on short-form video (e.g., Kinetics 5s clips). \KAcamera{HERO \cite{hero} uses a hierarchical loss between video clips (few seconds long) and their frames using only clip-level text, while %
\cite{cmhse} enhances parent-level understanding for video-to-para retrieval and action recognition by concatenating text sentences to form (non-abstractive) paragraphs for hierarchical training.}

All these methods only focus on localized narrations/captions. A single text sentence is matched to a clip that is typically a few seconds long.
There are two reasons for choosing smaller temporal windows: a) the narrations typically span only a few seconds, and b) longer clips introduce computational overload that makes training difficult.  
In contrast, we devise a hierarchical 
approach to use both clip-level narrations spanning a few seconds and abstractive video-level summaries spanning several minutes. %
We show that clip feature aggregation  makes %
learning computationally feasible, and  that using such hierarchical text descriptions improve both clip-level and video-level tasks.

\begin{figure*}[t]
\centering
\includegraphics[width=\textwidth]{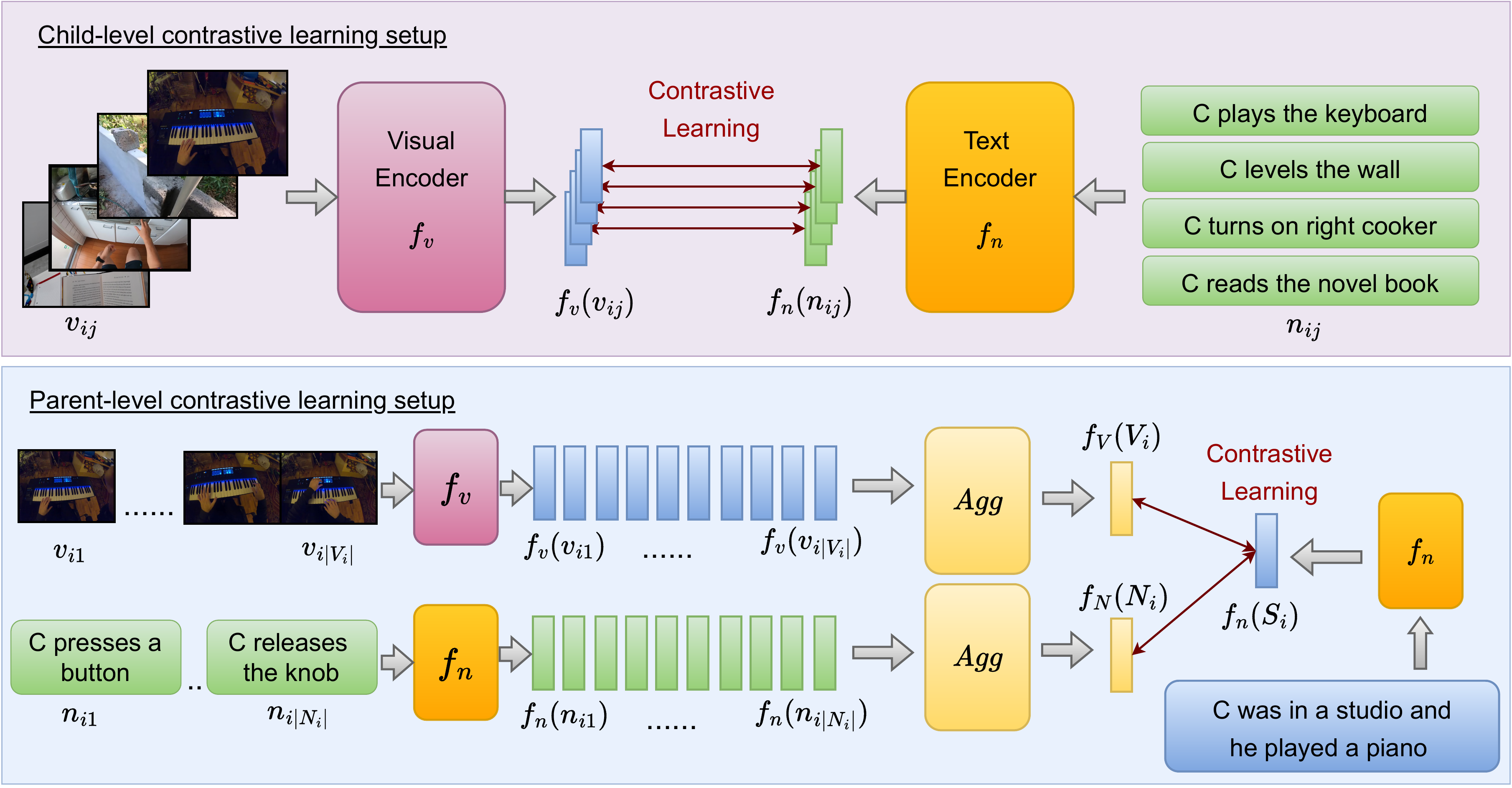}
\caption{Schematic representation of our proposed approach. In the clip-level contrastive learning setup (top), we match video clips with their corresponding narrations. %
The selected clips in one batch are from different videos, as shown. In our novel parent-level contrastive learning setup (bottom), we sample short-term features and aggregate them into a long-term representation followed by contrastive matching with the summary feature. %
These clips are sampled from the same video. Note that $f_v$ and $f_n$ are common in both stages, and also trainable in both. (For simplicity, figure only shows positive pairs in the contrastive setup.)}
\label{fig:method}
\vspace{-0.10in}
\end{figure*}

\section{Technical Approach}
\label{sec:method}

We propose \modelname, a novel video-language model that captures both clip- %
and video-level relations. \cref{fig:method} overviews our method. Next, we describe the annotations (\cref{sec:video-annotations}), formalize the embedding learning approach (\cref{sec:joint-video-text}), and discuss the feature aggregation strategy (\cref{sec:clip-agg}).  
Finally, we describe the loss function (\cref{contrastive}), 
training process (\cref{pre-train-strategy}), and implementation details (\cref{sec:impl}).

\subsection{Hierarchical video annotations} 
\label{sec:video-annotations}

Consider a hierarchically annotated video dataset, $\mathcal{D}_L~=~\{ (V_i, N_i, S_i) \}_{i=1}^{|\mathcal{D}_L|}$ where $V_i$ is a long video, $N_i$ is a sequence of text narrations describing every atomic action in the video, and $S_i$ is a high-level text summary for the whole video. Notationally, $V_i~=~\{v_{ij}\}_{j=1}^{|V_i|}$ is an ordered collection of short clips $v$ (each spanning a few seconds) and $N_i~=~\{n_{ij}\}_{j=1}^{|N_i|}$ is an ordered collection of narrations $n$. 
Note that there is no constraint on the temporal span of the video %
$V_i$, but in our experiments they are
typically in minutes. As an illustration, $n_{ij}$ can be \textit{``he cleans the painting brush"} or \textit{``he rubs the excess paint"} whereas high-level summary $S_i$ will be \textit{``he was painting in a drawing room"}. The clip $v_{ij}$ contains a visual demonstration of the narration $n_{ij}$, whereas $S_i$ is an abstractive summary of the full video $V_i$. 
The idea is for clip-level representations to capture fine-grained actions in a video, while video-level representations should capture the overall goal of the task.

We leverage the Ego4D dataset~\cite{ego4d,ego4dcons} for training our model.  Ego4D consists of 3,670 hours of wearable camera video of daily-life activity, as captured by 931 unique camera wearers around the world.  Among the Ego4D annotations are text descriptions (``narrations") of every action performed by the camera wearer, as well as video-level text summaries, which meet our requirements for $N$ and $S$, respectively.  The free-form narrations are written at timepoints selected by the annotators to capture every action performed.  Specifically, annotators first watched a full 5-minute video and wrote a short 1-3 sentence summary for the overall activity and environment. 
Then annotators were asked to pretend they were describing everything occurring in the video to a friend on the phone who cannot see the video.  The result is a temporally dense play-by-play description---13.2 sentences per minute on average, for a total of 3.85M sentences (see Appendix D in~\cite{ego4d} for details).

\subsection{Hierarchical joint video and text embedding}
\label{sec:joint-video-text}

In our hierarchical setup, we have short-term video segment $v$ and short-term text $n$.
We want to learn short-term representations $f_v(v)$ and $f_n(n)$, which we refer to as the visual short-term features and the textual short-term features. At the long-term level, we have $V$ and $N$ as a collection of multiple $v$ and multiple $n$, respectively. Simultaneously, we want to learn long-term representations $f_V(V)$ and $f_N(N)$ (referred to as long-term visual feature and long-term text feature, respectively). Finally, we have $f_n(S)$, the long-term summary feature, which is typically a few sentences long and hence is also encoded with $f_n$. 

The goal is to project $v, n, V, N, S$ into a common space 
such that semantically related features are close. %
Mathematically, for any suitably selected similarity metric $sim()$ and $\forall i_1, i_2, j_1, j_2 \text{~such that~} (i_1,j_1)~\neq~(i_2,j_2)$, we would like to fulfill a {\em child-level} matching constraint: 
\begin{equation}
\label{eq:clip_match}
  sim\left( f_v(v_{i_1j_1}), f_n(n_{i_1j_1})\right)~>~ sim(f_v(v_{i_1j_1}), f_n(n_{i_1j_2}))  
\end{equation}
and $\forall i, j \text{~such that~} i \neq j$, as well as {\em parent-level} matching constraints:
\begin{align}
    sim(f_V(V_i), f_n(S_i))~&>~sim(f_V(V_i), f_n(S_j)) \label{eq:vs_match}\\
    sim(f_N(N_i), f_n(S_i))~&>~sim(f_N(N_i), f_n(S_j)).\label{eq:ns_match}
\end{align}
Overall, Eq.~\ref{eq:clip_match} implies corresponding short-term representations should have higher similarity than non-matching ones, Eq.~\ref{eq:vs_match} (and Eq.~\ref{eq:ns_match}) implies a video (and narrations) should have a higher similarity with its summary than with other summaries. Note that since we project both short-term and long-term features into a common space, we are allowing features even at different hierarchical levels to come close in the embedding space if they are semantically similar.

\subsection{Efficient long-term features via aggregation}
\label{sec:clip-agg}

Obtaining long-term features is challenging in both visual and text modalities. Directly computing a long-term visual feature requires more resources due to its large video size and often leads to inferior performance and memory overflows~\cite{memvit,ltfb,clip-hitchhiker,scsampler}. Self-attention models are suitable architectures for capturing long-term dependencies, but they are challenging to apply to large collections of text sentences (e.g., long documents)
due to quadratic dependence on the token sequence length in transformer models \cite{bert}. Longformer \cite{longformer} mitigates this problem by multi-level global and local attentions. 

Taking inspiration from these works in both visual and textual domains, we use aggregations of short-term features as long-term representations $f_V$ and $f_N$. Following this strategy, we define the long-term visual representation $f_V$ as $f_V(V_i) = Agg\left(\{f_v(v_{ij})\}_{j=1}^{|V_i|}\right)$. Similarly, the long-term textual representation $f_N$ is defined as $f_N(N_i) = Agg\left(\{f_n(n_{ij})\}_{j=1}^{|N_i|}\right)$. %
We consider two aggregator functions $Agg(.)$. The first uses a self-attention transformer block in order to capture long-term dependencies over the entire video. We use positional encodings in order to provide the model with the ability to embed temporal order information in the video-level representation. We denote with \textbf{HierVL-SA} the variant of our model based on this self-attention aggregator. The second form of aggregation that we consider is simple average pooling (i.e., a parameter-free aggregator), which produces long-term features with equal contributions from all short-term features. This aggregator does not preserve order information. We name his version \textbf{HierVL-Avg}. %
We use the same aggregator in both modalities since $f(v)$ and $f(n)$ have the same dimensions (and, in fact, equal values for matching visual-text pairs in an ideal contrastive training).

\subsection{Contrastive pretraining objective}
\label{contrastive}

As introduced previously, we learn the representations at two levels---child-level $f_v, f_n$ and parent-level $f_V, f_N$. For child level representations, the pretraining objective is similar to prior work \cite{howto100m,egovlp,mil-nce,videoclip} that relates short-term visual representations to short-term textual representations. In particular, we use a variant of EgoNCE \cite{egovlp}, an action- and scene-aware variation of InfoNCE \cite{infonce}. EgoNCE groups similar actions as positives and temporally close distinct actions as hard negatives. In contrast, we omit the latter, %
since our hierarchical setup \emph{ought} to bring together distinct actions with the same camera-wearer intent. 
  Overall, the short-term pretraining objective is:
$$\mathcal{L}_{child} = \frac{1}{|\Tilde{\mathcal{B}}|} \sum_{i \in \tilde{\mathcal{B}}} \log \left(\frac{\sum_{j \in \mathcal{\tilde{P}}_i} \text{exp}({f_v(v_i)^Tf_n(n_j)})}{ \sum_{j \in \tilde{\mathcal{B}}} \text{exp}(f_v(v_i)^Tf_n(n_j))}\right)$$
where $\mathcal{\tilde{B}}$ is the overall set of short-term features and $\mathcal{\tilde{P}}$ is the per-instance set of action-aware positive samples (see \cite{egovlp} for details).  See~\cref{fig:method} (top). 

At the parent level, we use a similar pretraining objective between $S$-$V$ and $S$-$N$. See~\cref{fig:method} (bottom).  As discussed in \cref{sec:clip-agg}, we aggregate $v$ to obtain $V$ (and aggregate $n$ to get $N$). Since the short-term matching already contrasts $v$ and $n$, we do not contrast $f_V$ and $f_N$ again at the parent-level. 
Overall, the long-term pretraining objective is $\mathcal{L}_{parent} = \mathcal{L}^{SV}_{parent} + \mathcal{L}^{SN}_{parent}$ where 
$$\mathcal{L}^{SV}_{parent} = \frac{1}{|\Tilde{\mathcal{B}}|} \sum_{i \in \tilde{\mathcal{B}}} \log \left(\frac{\sum_{j \in \mathcal{\tilde{P}}_i} \text{exp}({f_V(V_i)^Tf_n(S_j)})}{ \sum_{j \in \tilde{\mathcal{B}}} \text{exp}(f_V(V_i)^Tf_n(S_j))}\right)$$
and similarly for $\mathcal{L}^{SN}_{parent}$. For the parent-level feature, negatives for a summary text $S_i$ are both visual and textual representations chosen from outside the temporal span of $S_i$.

\subsection{Training strategy} 
\label{pre-train-strategy}

So far, we discussed our approach for hierarchical video-language pretraining. To realize this setup, we employ a joint training approach. First, we train $m$ batches of short-term visual and textual pairs $(v, n)$ --- thus training $f_v$ and $f_n$. Subsequently, we train one batch of long-term features --- thereby training $f_V$ and $f_N$. Recall that $f_V(.) = Agg(f_v(.))$ and $f_N(.) = Agg(f_n(.))$. Therefore, in this batch, we update the weights of $Agg$ as well as short-term $f_v$ and $f_n$. \ccCR{The contrastive objective is detailed in \cref{contrastive}.}

The motivation behind training both levels of annotations together is to ensure the functions $f_v$ and $f_n$ optimize for both short-term and long-term features, i.e., both are influenced by the text summaries. Other alternatives are (a) using separate models for clip-level and video-level features, but that increases the parameters in the model and makes the training difficult (both in terms of convergence and GPU usage), and (b) training with only clip-level data and fine-tuning it for video-level (or vice-versa), but %
such strategies are known to
lead to catastrophic forgetting \cite{catastrophic-forgetting-1,catastrophic-forgetting-2,catastrophic-forgetting-3}.

\begin{figure}[t]
\centering
\includegraphics[width=0.47\textwidth]{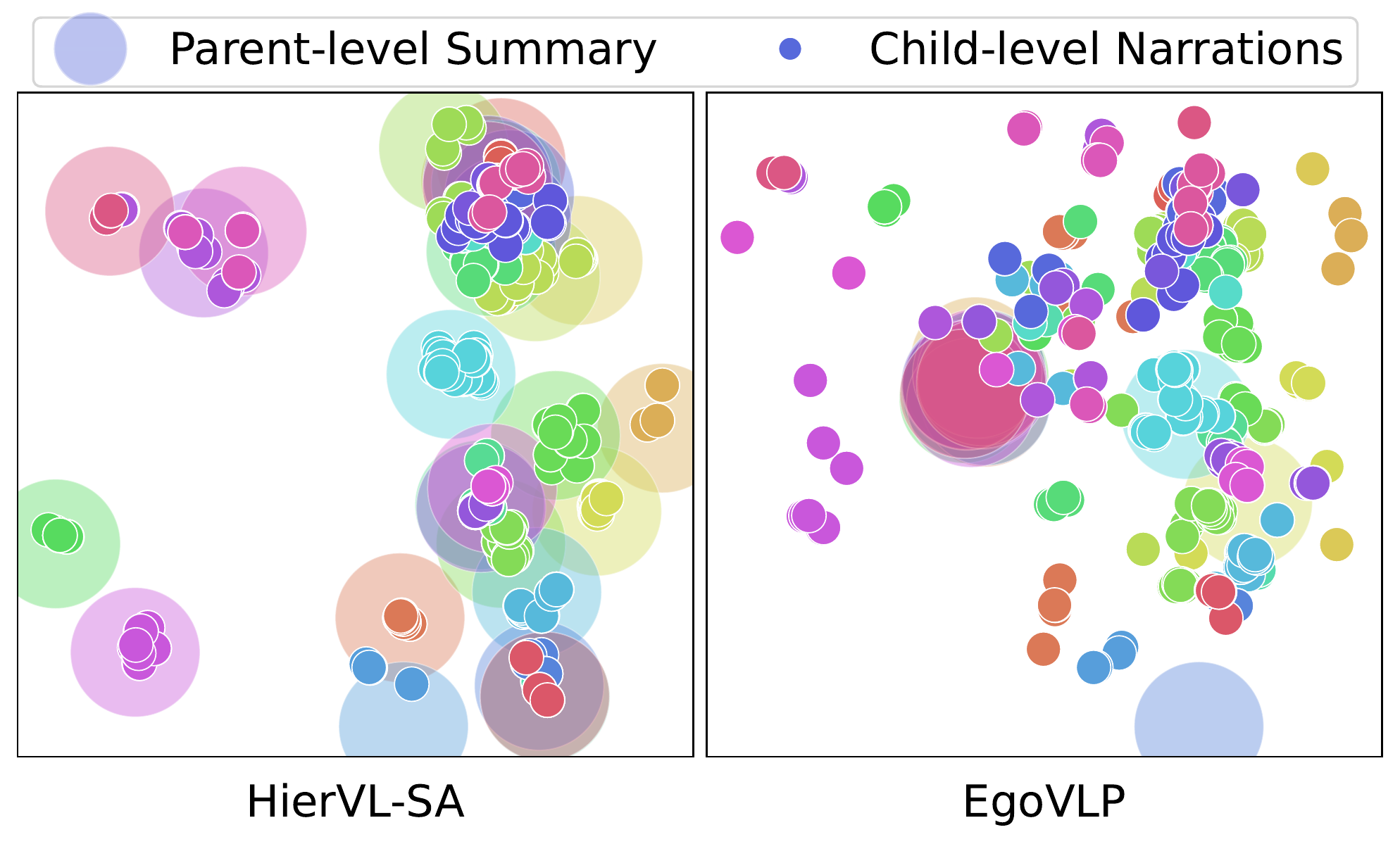}
\caption{T-SNE plot of learned features from our \modelname-SA (left) and EgoVLP \cite{egovlp} (right).  See text and Supp.~for details.}

\label{fig:tsne}
\vspace{-0.10in}
\end{figure}

\cref{fig:tsne} visualizes the learned features for 500 summary texts and their child narrations using our $f_n$ (left) and EgoVLP's features (right).  
While summary features in EgoVLP are unrelated to the  narrations, 
\modelname~captures their natural hierarchy, as seen by the colors clustering together in the embedding space. 
This reshaping of the features reflects how our clip-level features convey context about the higher-level intent of the camera wearer.

\subsection{Implementation Details}
\label{sec:impl}

\noindent \textbf{Network architecture.} To learn the video feature extractor $f_v$, we use a standard FrozenInTime \cite{frozenintime} video backbone, which is a slight deviation from TimeSformer \cite{timesformer} and inspired from ViT \cite{vit}. ViT-based vision transformers are frequently used as a feature extractor \cite{egovlp,clip} owing to their superior performance compared to other backbones. The video representation $f_v$ is learned from scratch; the output representation is the output of the final CLS token. We choose frames at 1 fps for short-term clips.
Next, the text feature extractor $f_n$ is a DistillBERT \cite{distilbert} architecture which achieves performance on-par with BERT \cite{bert} but offers the benefit of being lighter. %

\noindent \textbf{Aggregator.}  Our HierVL-SA variant is implemented by means of a 6-layer self-attention block of the TimeSformer architecture \cite{timesformer} and \modelname-Avg is averaging of features. In order to have a constant batch size, for both HierVL-SA and HierVL-Avg, we aggregate 16 short-term representations uniformly sampled from the entire video.

\noindent \textbf{Training setup and parameters.} We pretrain our architecture on 4 nodes, each with eight 32 GB NVIDIA V100 GPUs for 10 epochs for two days. We use AdamW \cite{adamw} optimizer with a learning rate of $3\times10^{-5}$. We train one batch of video-level aggregation after every $m=5$ epoch of clip-level training. 
We use a batch size of 16 per GPU for short-term contrastive learning %
and 1 per GPU for long-term video-level contrastive learning. Recall that one video-level batch consists of 16 clips of the same video.

\section{Experiments}
\label{sec:exp}

We first pretrain our architecture \ccCR{with the setup and parameters discussed in \cref{sec:impl}} and report its results on multiple tasks aimed directly at gauging the quality of the learned video features (Sec.~\ref{sec:exp-evaluation}).  
Next, we show that our pretrained model improves the state of the art on a variety of downstream tasks covering both short- and long-term understanding (Sec.~\ref{sec:exp-downstream}). %

{
 \setlength{\tabcolsep}{1pt}
 \setlength{\extrarowheight}{1.5pt}
\begin{table*}[t]\footnotesize
\begin{center}
\begin{tabular}{ L{3.2cm}||C{1.0cm}|C{1.0cm}|C{1.0cm}|C{2.2cm}||C{1.9cm}|C{1.9cm}||C{1.9cm}|C{1.9cm}  }

 \hline
&Joint&&&& & & \multicolumn{2}{c}{EgoMCQ} \\
\cline{8-9}
 Method&train&Hier&Summ&Aggregation& Summ MCQ&Shuffle MCQ&Inter-video &Intra-video\\
 \hline
 \textcolor{gray}{EgoVLP \cite{egovlp}}   &&&&& \textcolor{gray}{---}    & \textcolor{gray}{---} &\textcolor{gray}{90.6}   & \textcolor{gray}{57.2}\\
  EgoVLP (reproduced)   &---& \xmark & \xmark & \xmark&  89.0    & 20.0 & 90.1   & \textbf{54.0} \\
 \rowcolor{Gray}
 \modelname-Avg (Ours) &\cmark&\cmark&\cmark&Average& \underline{95.2} & 20.0 & 90.3 & \underline{53.1}\\
 \rowcolor{Gray}
 \modelname-SA (Ours)&\cmark&\cmark&\cmark&Self-attention& \textbf{95.4} & \textbf{26.8} & 90.5 & 52.4\\
\hline  
 \modelname-w/o Joint &\xmark&\xmark&\cmark&\xmark& 89.8 & 24.2 & 72.0 &  29.4\\
  \modelname-w/o Hier &\cmark&\xmark&\cmark&\xmark& 93.7 & 20.0 & \underline{90.7} &  50.5\\
  \modelname-w/o Summ&\cmark&\cmark&\xmark&Self-attention& 20.0   & 22.1 & \textbf{90.8} & 52.1\\
 \modelname-w/o Summ $\leftrightarrow$ Narr &\cmark&\cmark&\cmark&Self-attention& 94.7   & \underline{26.1} & 90.4 & 50.0\\
 \hline
\end{tabular}
\end{center}
\vspace{-0.10in}
\caption{Pretraining accuracy on EgoMCQ, SummaryMCQ, and ShuffleMCQ on Ego4D pretraining, compared to EgoVLP (top) and ablations. For all validation sets, chance corresponds to 20.0 accuracy. Our proposed method using both hierarchy and long-term summary performs better than all baselines on the long-term SummaryMCQ and ShufleMCQ tasks. As expected, both methods are comparable in the short-term MCQ task. ---: N/A, bold is best, underline is second best.
}
\label{tab:pre-train}
\end{table*}

}

\subsection{Pretraining Evaluation}
\label{sec:exp-evaluation}

We use Ego4D \cite{ego4d,ego4dcons} for our contrastive pretraining. Ego4D has two-level hierarchical annotations---short-term step-by-step narrations and a long-term summary of the demonstration as observed by an annotator. We maintain the same training and validation split as in \cite{egovlp}. Overall, there are $3.8$M short-term narrations and $120$K long-term summary annotations. %

\noindent \textbf{Pretraining evaluation tasks.}
We evaluate the quality of pretraining on three tasks defined on the Ego4D dataset: EgoMCQ (\textbf{m}ultiple-\textbf{c}hoice-\textbf{q}uestion, introduced in EgoVLP \cite{egovlp}), as well as two new benchmarks that we propose --- SummaryMCQ and ShuffleMCQ.  In \textbf{EgoMCQ}, the  model is given a narration prompt along with five candidate clips and must match the prompt with the correct video clip, with accuracy as the performance metric. Intra-video and Inter-video are two splits of the validation data where the candidate video clips are selected from the same or the other videos, respectively. \textbf{SummaryMCQ} mimics the video-language matching test of EgoMCQ but here the model is given a \emph{summary} and five candidate \emph{long-term} video options. The options are videos spanning the whole summary duration. While EgoMCQ validates clip-level performance, SummaryMCQ validates video-level performance. Finally, \textbf{ShuffleMCQ} is designed to evaluate temporal understanding: a summary text is given, and only the correct option maintains the temporal order among clips. The other four video options are generated by randomly reshuffling clips of the original video.

\noindent \textbf{Comparison to EgoVLP.} Our main comparison is to EgoVLP \cite{egovlp}, since our model adopts the same architecture and uses its EgoNCE as the short-term loss in the objective.
However, while our method leverages a hierarchical contrastive training that makes use of summary information, EgoVLP only focuses on short-term visual-textual correspondences. For SummaryMCQ, we use parameter-free averaging to compute the aggregate representation. %

Table \ref{tab:pre-train} shows the results.\footnote{The first row corresponds to the numbers reported in EgoVLP \cite{egovlp} and the second row corresponds to the numbers that we reproduced using the same codebase. We attribute the difference in performance to different hardware configurations.} %
EgoVLP~\cite{egovlp} and both variants of our \modelname~perform similarly on EgoMCQ, consistent with the fact this task requires short-term information only. In contrast, \modelname-SA obtains significantly better accuracy on the video-level (long-term) tasks, SummaryMCQ and ShuffleMCQ. Specifically, \modelname-SA outperforms EgoVLP by more than $6\%$  on SummaryMCQ. This highlights our model's ability to capture long-term intent more effectively than the aggregated short-term features of EgoVLP. On ShuffleMCQ, both EgoVLP and \modelname-Avg are no better than chance (20\%). This reflects how neither model captures the temporal order information that is essential to distinguish between the original summary and shuffled videos. Conversely, \modelname-SA exhibits stronger performance, producing a gain of 6.8\% over these models (a relative gain of 34\%).   In short, our hierarchical learning shines for the long-term video tasks, successfully encoding the longer-term dependencies between events. \KAcamera{We also observe that HierVL-SA outperforms EgoVLP with varying model sizes. Thus, further scaling models would not diminish the need for our architecture (see Supp).}

\noindent \textbf{Ablating design choices.} 
The bottom portion of  Table \ref{tab:pre-train} includes several variants of our \modelname, in order to ablate the different design choices. %
Our proposed architecture has three distinct components: (a) a hierarchical model that operates at two levels (parent-level summaries and child-level narrations), (b) use of text summaries as a supervision, and (c) the joint training of these hierarchical annotations. 

\textbf{\modelname-w/o~Joint} is a variant used to investigate the effectiveness of joint training (component c). We start \modelname-w/o Joint  with EgoVLP pretrained weights and train the whole network ($f_v, f_n, Agg$) using summaries only, i.e., {\em without} narrations. In this variant, the clip representations are indirectly supervised by means of the parent loss. %
We can see that while \modelname-w/o~Joint achieves decent results on the two video-level tasks, its performance on EgoMCQ is much lower than that achieved by EgoVLP, which is its initialization. This suggests that summaries by themselves are not sufficient to supervise the learning of strong clip-level representations. 

\begin{figure}[t]
\centering
\includegraphics[width=0.47\textwidth]{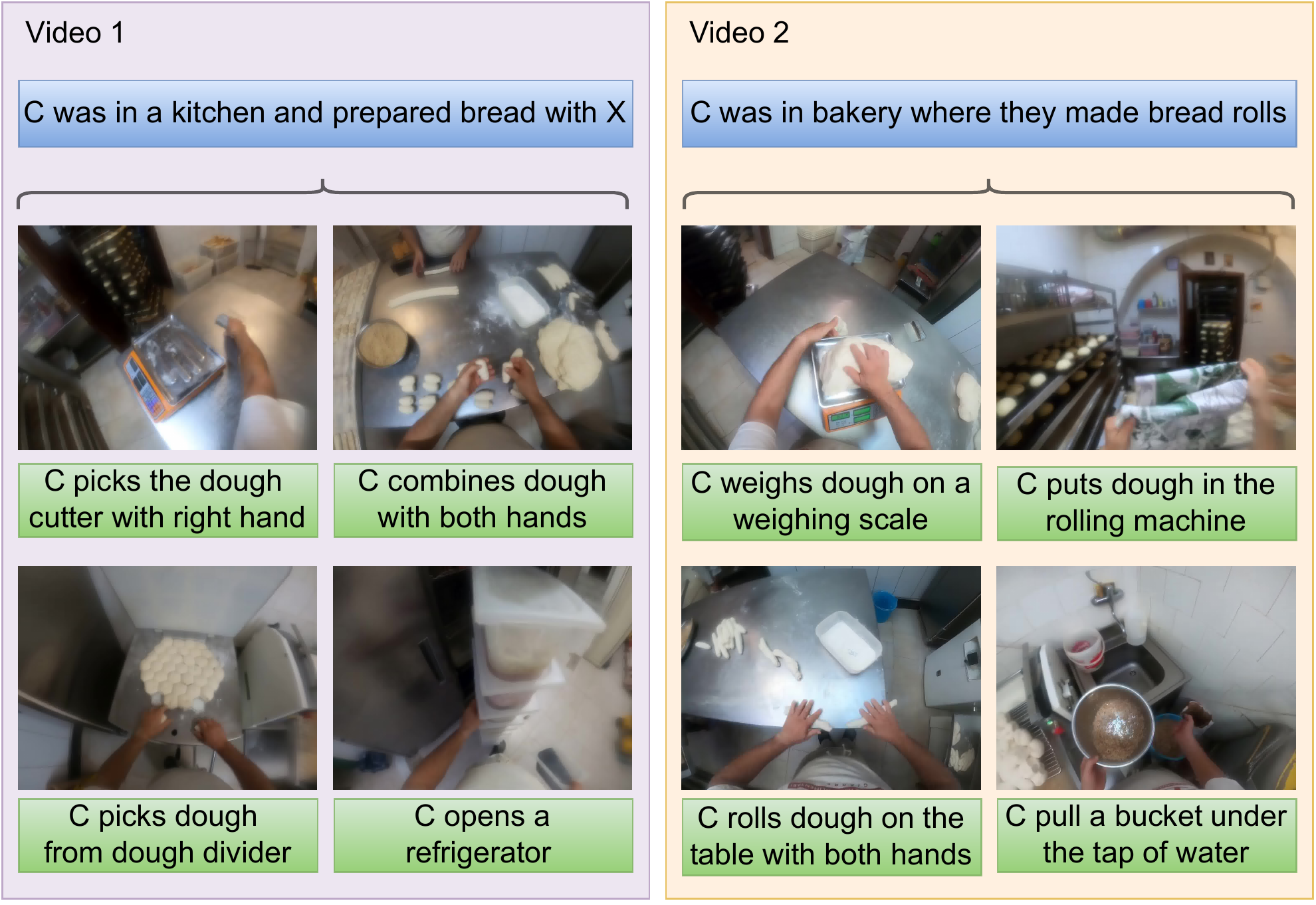}
\vspace*{-0.12in}
\caption{Examples of video segments that are close in the embedding space despite coming from different videos and representing different short-term steps. Both the videos have the same high-level objective, i.e. making bread.}

\label{fig:qual}
\vspace{-0.10in}
\end{figure}

\textbf{\modelname-w/o Hier} uses (b, c) but not (a), i.e., we use summary supervision without a hierarchical model. We randomly assign the summary text annotation to one of the short-term  segments. Importantly, this baseline uses the same amount of supervision as our proposed \modelname, yet it has overall lower performance %
(except for a marginal gain on EgoMCQ Inter-video).  This highlights the effectiveness of our hierarchical training scheme. 

\textbf{\modelname-w/o Summ} uses (a, c) but not (b), i.e., the supervision does not come from the summary text. \KAcamera{Note, this represents the main idea from~\cite{cmhse}.}  The parent-level positives for contrastive learning are $f_V$ and $f_N$. The objective of this ablation is to determine if high-level summaries are needed, or whether an aggregation of narrations can serve as a high-level representation. We observe that this variant is considerably less effective than \modelname-SA on the two video-level tasks of SummaryMCQ and ShuffleMCQ. This is an important result, as it suggests that the high-level intent expressed by the human annotator in the summary is effectively captured by \modelname-SA and this human supervision cannot be adequately replaced by an aggregation of short-term narrations. 

Finally, \textbf{\modelname-w/o Summ$\leftrightarrow$Narr} investigates the need for an additional text-only parent-level matching, as given in \cref{eq:ns_match}. This ablation checks the effect of only matching $f_V(V) \leftrightarrow f_n(S)$ vs.~matching both $f_V(V) \leftrightarrow f_n(S)$ and $f_N(N) \leftrightarrow f_n(S)$. We see that imposing additional supervision between child and parent text features does increase the performance on all validation sets. %

\subsection{Downstream Evaluation}
\label{sec:exp-downstream}

We evaluate the representation learned by \modelname{} on  multiple downstream tasks.
 
\vspace*{-0.15in}
\paragraph{Datasets.}  In addition to \textbf{Ego4D}~\cite{ego4d} we use 
\textbf{Charades-Ego} \cite{charades-ego}, which consists of 7,860 videos recorded from both first and third person viewpoints, with 157 action classes; %
\textbf{EPIC-Kitchens-100} \cite{epic-kitchens-100,epic-ijcv}, an egocentric video of 100 hours of unscripted activities in 45 home kitchens in 4 cities; and \textbf{HowTo100M} \cite{howto100m}, a large-scale YouTube dataset covering $23$K visual ``how-to" tasks. %

\paragraph{Downstream tasks.} We consider the following tasks:
\begin{itemize}
\item \textbf{Long-Term Anticipation (LTA).} Ego4D's LTA challenge requires the model to predict the next 20 actions given the current action (verb, noun). %
Metric is Edit Distance (ED) \cite{ego4d}. %

\item \textbf{Action Recognition.} Charades-Ego's task requires predicting the action among 157 categories. Metric is mAP (mean average precision). We evaluate both the zero-shot and fine-tuned settings.

\item \textbf{Multi-Instance Retrieval (MIR).} EPIC-Kitchens-100's MIR is a text-to-video and video-to-text retrieval task. %
Metrics are mAP and nDCG (normalized Discounted Cumulative Gain) for both V$\rightarrow$T and T$\rightarrow$V. We report their averages. %
Again, we evaluate in both zero-shot and fine-tuned settings.

\item \textbf{Video Classification.} To demonstrate the transfer ability of our pretraining, we perform linear probing on the most frequent 100 classes in HowTo100M. %
Metric is classification accuracy.
\end{itemize}

Throughout, we report relevant comparisons from the best existing methods in the literature, as well as the ``w/o Hier" ablation, which uses the exact same summary data/supervision as \modelname, hence pinpointing the influence of our hierarchical training idea.

{
 \setlength{\tabcolsep}{1pt}
 \setlength{\extrarowheight}{1.5pt}
\begin{table}[t]\footnotesize
\begin{center}
\begin{tabular}{ L{3.0cm}||C{1.6cm}|C{1.6cm}|C{1.6cm}  }

 \hline
 Method& Verb ED $\downarrow$ & Noun ED $\downarrow$ & Act. ED $\downarrow$ \\
 \hline
 Ego4D baseline \cite{ego4d}   &  0.7389   & 0.7800 & 0.9432 \\
  Robovision \cite{lta-srijan}   &  0.7389   & 0.7688 & 0.9412 \\
  I-CVAE \cite{intention}   &  0.7526   & 0.7489 & \underline{0.9308} \\
 \modelname-w/o Hier & 0.7691 & \underline{0.7454} & 0.9451\\
\modelname-Avg (Ours) & \textbf{0.7223} & 0.7527 & 0.9401 \\
 \modelname-SA ~(Ours)& \underline{0.7239} & \textbf{0.7349} & \textbf{0.9275} \\
 \hline
\end{tabular}

\end{center}
\vspace{-0.2in}
\caption{Errors on Ego4D Long Term Anticipation (LTA) Challenge. ED is the edit distance at $Z=20$, lower the better.}
\label{tab:lta}
\vspace{-3mm}
\end{table}

}

{
 \setlength{\tabcolsep}{1pt}
 \setlength{\extrarowheight}{1.5pt}
\begin{table}[t]\footnotesize
\begin{center}
\begin{tabular}{ L{3.0cm}||C{2.2cm}|C{2.2cm}  }

 \multicolumn{3}{c}{Zero-shot}  \\
 \hline
 Method& Task ckpt mAP& PT ckpt mAP\\
 \hline
 EgoVLP \cite{egovlp}   &  25.0   & 19.4 \\
  \modelname-w/o Hier    &   24.6 & \underline{24.5} \\
  \modelname-Avg (Ours)   &  \underline{25.2} & 23.9   \\
\modelname-SA (Ours)& \textbf{26.0} & \textbf{25.0} \\
 \hline
\end{tabular}

\begin{tabular}{ L{3.5cm}||C{2.1cm}  }

 \multicolumn{2}{c}{Fine-tuned}  \\
 \hline
 Method&mAP\\
 \hline
 Actor \cite{actor} &   20.0 \\
 SSDA \cite{charades-comparison} & 23.1\\
 I3D \cite{charades-comparison}  & 25.8 \\
 Ego-Exo \cite{ego-exo} & 30.1\\
 EgoVLP \cite{egovlp}  & 32.1\\
   \modelname-w/o Hier   &  \underline{32.6}  \\
  \modelname-Avg (Ours)   &  \underline{32.6}  \\
 \modelname-SA (Ours) & \textbf{33.8}\\
 \hline
\end{tabular}
\end{center}
\vspace{-0.2in}
\caption{Zero-shot (top) and fine-tuned (bottom) accuracy on Charades-Ego  action recognition. %
We outperform EgoVLP and resist overfitting in the zero-shot case.  Our fine-tuned performance is the best reported in the literature to-date for this dataset. }
\label{tab:charades-ego}
 \vspace{-3mm}
\end{table}

}
{
 \setlength{\tabcolsep}{1pt}
 \setlength{\extrarowheight}{1.5pt}
\begin{table}[t]\footnotesize
\begin{center}
\begin{tabular}{ L{3.5cm}||C{2.2cm}|C{2.2cm}  }

 \multicolumn{3}{c}{Zero-shot}  \\
 \hline
 Method& mAP Avg& nDCG Avg\\
 \hline
 EgoVLP \cite{egovlp}   &  16.6   & 23.1 \\
 \modelname-w/o Hier   &  \underline{17.8}   & \underline{24.1} \\
 \modelname-Avg (Ours) & 16.7 & 23.5\\
 \modelname-SA (Ours)& \textbf{18.9} & \textbf{24.7}\\
 \hline
\end{tabular}

\begin{tabular}{ L{3.5cm}||C{2.2cm}|C{2.2cm}  } 

 \multicolumn{3}{c}{Fine-tuned}  \\
 \hline
 Method& mAP Avg& nDCG Avg\\
 \hline
 MI-MM w/ S3D \cite{s3d-g} & 29.2 & 44.7 \\
  MME \cite{wray2019fine} w/ TBN \cite{epic-fusion} & 38.5 & 48.5 \\
   JPoSE \cite{wray2019fine} w/ TBN \cite{epic-fusion} & 44.0 & 53.5 \\
 EgoVLP \cite{egovlp}   &  \underline{45.0}   & 59.4 \\
   \modelname-w/o Hier   &  44.7   & \underline{59.8} \\ %
 \modelname-Avg (Ours) & 44.9 & \underline{59.8}\\
 \modelname-SA (Ours)& \textbf{46.7}  & \textbf{61.1}\\
 \hline
\end{tabular}
\end{center}
\vspace{-0.2in}
\caption{Zero-shot and fine-tuned performance on EPIC-Kitchens-100 dataset for multi-instance retrieval task.}
\label{tab:epic-mir}
\vspace{-3mm}
\end{table}

}
{
 \setlength{\tabcolsep}{1pt}
 \setlength{\extrarowheight}{1.5pt}
\begin{table}[t]\footnotesize
\begin{center}

\begin{tabular}{ L{3.5cm}||C{2.1cm}|C{2.1cm}  }
 \hline
 Method&Inference&Accuracy\\
 \hline
 EgoVLP \cite{egovlp} & Avg &  53.4 \\
 \modelname-SA (Ours) & Self-attention &54.6\\
 \modelname-Avg (Ours) & Avg &\underline{63.3}\\
  \modelname-SA (Ours) & Avg &\textbf{64.6}\\
 \hline
\end{tabular}
\end{center}
\vspace{-0.2in}
\caption{Linear probe results on HowTo100M video classification.}
\label{tab:howto100m-vc}
\vspace{-3mm}
\end{table}

}

\vspace*{0.15in}
\textbf{Ego4D LTA:} \cref{tab:lta} shows results on the test set of Ego4D LTA challenge. 
The models need to forecast the future $20$ actions, which is non-trivial even for humans. We improve the state of the art in both verb and noun predictions. %
Additionally, ours is the best performing method on the public leaderboard at the time of submission  (in \cref{tab:lta} we only compare with published works).    \modelname-w/o Hier does not perform well despite also having access to the summaries, thus asserting the effectiveness of our hierarchical training. 
We use our learned representations $f_v$ and $Agg$ followed by a multi-headed decoder, as in the baseline \cite{ego4d}.
This result shows the effectiveness of both our learned feature aggregator (long-term) as well as short-term visual encoder $f_v$.

\textbf{Charades-Ego Action Recognition.}
 \cref{tab:charades-ego} (top) shows the zero-shot results.
EgoVLP \cite{egovlp} reports overfitting when transferring from Ego4D to Charades-Ego and hence chooses another pretraining checkpoint. There is a significant gap in the performance between the two checkpoints. We report results on both---best performing pretraining checkpoint (denoted as PT ckpt) and the checkpoint chosen by EgoVLP (denoted as Task ckpt). Our model does not overfit when transferring to Charades-Ego; our performance on the corresponding checkpoints are $5.6\%$ and $1.0\%$ higher. In this downstream evaluation, only the short-term visual encoder $f_v$ (frozen) is required. Clearly, our hierarchical pretraining improves short-term features as well.

\cref{tab:charades-ego} (bottom) shows the fine-tuned results for the same task.  Here, to compare against state-of-the-art methods, we fine-tune the model starting from our best pretrained checkpoint (having $25.0\%$ mAP for \modelname-SA). We outperform the current state-of-the-art EgoVLP \cite{egovlp}. %
We fine-tune $f_v$ for this task, showing improvement in the short-term features.  To our knowledge, ours is the best reported result for this dataset in the literature. 

\textbf{EPIC-Kitchens-100 Multi-Instance Retrieval.} 
\cref{tab:epic-mir} (top) shows the zero-shot results.
We observe a gain of $2.3\%$ mAP and $1.6\%$ increase between the best method and our \modelname-SA. Our \modelname-Avg is also slightly better than the state-of-the-art method. In this task, we use both the short-term encoders $f_v$ and $f_n$ (both frozen) and thus this experiment also validates our claim of improved short-term representations via hierarchical learning.  \cref{tab:epic-mir} (bottom) shows our fine-tuning results for the same task. 
We fine-tune both $f_v$ and $f_n$.
We increase both metrics  %
compared to the state-of-the-art. %

\textbf{HowTo100M Video Classification.} \cref{tab:howto100m-vc} %
shows the results.  In this linear probe setting, all of $f_v, f_n$ and $Agg$ are frozen and only one additional linear layer is trainable (trainable parameters $25.7$K). We see that all of our learned representations are better than the baseline EgoVLP. Parameter-free averaging works well in video classification%
~\cite{clip-hitchhiker}. Therefore, we add a special case of \modelname-SA where we retain the pretrained $f_v$ and replace SA with average. This additional experiment also shows the superiority of short-term features $f_v$ in \modelname-SA compared to \modelname-Avg.

\section{Conclusion}
\label{sec:conclusion}

We introduce a novel hierarchical video-language embedding. Whereas current embeddings are oblivious to the long-term activity intent, %
\modelname~focuses on both short-term ``what is the person doing now" and long-term ``what the person aims to do". Through extensive experiments, we show that this improves both short-term and long-term video understanding.  Our model pushes the state-of-the-art on a variety of video challenges, including the overall best performance in the literature on Charades-Ego action recognition and Ego4D long-term anticipation.

\textbf{Acknowledgements:} We thank Ziad Al-Halah and Tushar Nagarajan for feedback on the manuscript.
KG is paid as a research scientist at Meta. UT Austin is supported in part by the IFML NSF AI Institute and NSF-CCRI.

{\small
\bibliographystyle{ieee_fullname}
\bibliography{egbib}

\begin{thebibliography}{100}\itemsep=-1pt

\bibitem{pre-computed-1}
Sami Abu-El-Haija, Nisarg Kothari, Joonseok Lee, Paul Natsev, George Toderici,
  Balakrishnan Varadarajan, and Sudheendra Vijayanarasimhan.
\newblock Youtube-8m: A large-scale video classification benchmark.
\newblock {\em arXiv preprint arXiv:1609.08675}, 2016.

\bibitem{whenwillyoudowhat}
Yazan Abu~Farha, Alexander Richard, and Juergen Gall.
\newblock When will you do what?-anticipating temporal occurrences of
  activities.
\newblock In {\em Proceedings of the IEEE conference on computer vision and
  pattern recognition}, pages 5343--5352, 2018.

\bibitem{frozenintime}
Max Bain, Arsha Nagrani, G{\"u}l Varol, and Andrew Zisserman.
\newblock Frozen in time: A joint video and image encoder for end-to-end
  retrieval.
\newblock In {\em Proceedings of the IEEE/CVF International Conference on
  Computer Vision}, pages 1728--1738, 2021.

\bibitem{clip-hitchhiker}
Max Bain, Arsha Nagrani, G{\"u}l Varol, and Andrew Zisserman.
\newblock A clip-hitchhiker's guide to long video retrieval.
\newblock {\em arXiv preprint arXiv:2205.08508}, 2022.

\bibitem{bansal2022my}
Siddhant Bansal, Chetan Arora, and CV Jawahar.
\newblock My view is the best view: Procedure learning from egocentric videos.
\newblock {\em arXiv preprint arXiv:2207.10883}, 2022.

\bibitem{longformer}
Iz Beltagy, Matthew~E Peters, and Arman Cohan.
\newblock Longformer: The long-document transformer.
\newblock {\em arXiv preprint arXiv:2004.05150}, 2020.

\bibitem{timesformer}
Gedas Bertasius, Heng Wang, and Lorenzo Torresani.
\newblock Is space-time attention all you need for video understanding?
\newblock In {\em ICML}, volume~2, page~4, 2021.

\bibitem{bi2021procedure}
Jing Bi, Jiebo Luo, and Chenliang Xu.
\newblock Procedure planning in instructional videos via contextual modeling
  and model-based policy learning.
\newblock In {\em Proceedings of the IEEE/CVF International Conference on
  Computer Vision}, pages 15611--15620, 2021.

\bibitem{activitynet}
Fabian Caba~Heilbron, Victor Escorcia, Bernard Ghanem, and Juan Carlos~Niebles.
\newblock Activitynet: A large-scale video benchmark for human activity
  understanding.
\newblock In {\em Proceedings of the ieee conference on computer vision and
  pattern recognition}, pages 961--970, 2015.

\bibitem{chang2020procedure}
Chien-Yi Chang, De-An Huang, Danfei Xu, Ehsan Adeli, Li Fei-Fei, and
  Juan~Carlos Niebles.
\newblock Procedure planning in instructional videos.
\newblock In {\em European Conference on Computer Vision}, pages 334--350.
  Springer, 2020.

\bibitem{vid-retrieval-1}
Xing Cheng, Hezheng Lin, Xiangyu Wu, Fan Yang, and Dong Shen.
\newblock Improving video-text retrieval by multi-stream corpus alignment and
  dual softmax loss.
\newblock {\em arXiv preprint arXiv:2109.04290}, 2021.

\bibitem{charades-comparison}
Jinwoo Choi, Gaurav Sharma, Manmohan Chandraker, and Jia-Bin Huang.
\newblock Unsupervised and semi-supervised domain adaptation for action
  recognition from drones.
\newblock In {\em Proceedings of the IEEE/CVF Winter Conference on Applications
  of Computer Vision}, pages 1717--1726, 2020.

\bibitem{ego4dcons}
Ego4D Consortium.
\newblock Egocentric live 4d perception ({Ego4D}) database: A large-scale
  first-person video database, supporting research in multi-modal machine
  perception for daily life activity.
\newblock {\em \url{https://sites.google.com/view/ego4d/home}}.

\bibitem{img-grounding-1}
Bo Dai, Yuqi Zhang, and Dahua Lin.
\newblock Detecting visual relationships with deep relational networks.
\newblock In {\em Proceedings of the IEEE conference on computer vision and
  Pattern recognition}, pages 3076--3086, 2017.

\bibitem{epic-ijcv}
Dima Damen, Hazel Doughty, Giovanni~Maria Farinella, , Antonino Furnari, Jian
  Ma, Evangelos Kazakos, Davide Moltisanti, Jonathan Munro, Toby Perrett, Will
  Price, and Michael Wray.
\newblock Rescaling egocentric vision: Collection, pipeline and challenges for
  epic-kitchens-100.
\newblock {\em International Journal of Computer Vision (IJCV)}, 130:33–55,
  2022.

\bibitem{epic-kitchens-100}
Dima Damen, Hazel Doughty, Giovanni~Maria Farinella, Antonino Furnari,
  Evangelos Kazakos, Jian Ma, Davide Moltisanti, Jonathan Munro, Toby Perrett,
  Will Price, et~al.
\newblock Rescaling egocentric vision: collection, pipeline and challenges for
  epic-kitchens-100.
\newblock {\em International Journal of Computer Vision}, 130(1):33--55, 2022.

\bibitem{lta-srijan}
Srijan Das and Michael~S Ryoo.
\newblock Video+ clip baseline for ego4d long-term action anticipation.
\newblock {\em arXiv preprint arXiv:2207.00579}, 2022.

\bibitem{bert}
Jacob Devlin, Ming-Wei Chang, Kenton Lee, and Kristina Toutanova.
\newblock Bert: Pre-training of deep bidirectional transformers for language
  understanding.
\newblock {\em arXiv preprint arXiv:1810.04805}, 2018.

\bibitem{img-retrieval-2}
Haiwen Diao, Ying Zhang, Lin Ma, and Huchuan Lu.
\newblock Similarity reasoning and filtration for image-text matching.
\newblock In {\em Proceedings of the AAAI Conference on Artificial
  Intelligence}, volume~35, pages 1218--1226, 2021.

\bibitem{pre-computed-2}
Jeffrey Donahue, Lisa Anne~Hendricks, Sergio Guadarrama, Marcus Rohrbach,
  Subhashini Venugopalan, Kate Saenko, and Trevor Darrell.
\newblock Long-term recurrent convolutional networks for visual recognition and
  description.
\newblock In {\em Proceedings of the IEEE conference on computer vision and
  pattern recognition}, pages 2625--2634, 2015.

\bibitem{img-grounding-2}
Qi Dong, Zhuowen Tu, Haofu Liao, Yuting Zhang, Vijay Mahadevan, and Stefano
  Soatto.
\newblock Visual relationship detection using part-and-sum transformers with
  composite queries.
\newblock In {\em Proceedings of the IEEE/CVF International Conference on
  Computer Vision}, pages 3550--3559, 2021.

\bibitem{vit}
Alexey Dosovitskiy, Lucas Beyer, Alexander Kolesnikov, Dirk Weissenborn,
  Xiaohua Zhai, Thomas Unterthiner, Mostafa Dehghani, Matthias Minderer, Georg
  Heigold, Sylvain Gelly, et~al.
\newblock An image is worth 16x16 words: Transformers for image recognition at
  scale.
\newblock {\em arXiv preprint arXiv:2010.11929}, 2020.

\bibitem{vid-retrieval-3}
Han Fang, Pengfei Xiong, Luhui Xu, and Yu Chen.
\newblock Clip2video: Mastering video-text retrieval via image clip.
\newblock {\em arXiv preprint arXiv:2106.11097}, 2021.

\bibitem{slowfast}
Christoph Feichtenhofer, Haoqi Fan, Jitendra Malik, and Kaiming He.
\newblock Slowfast networks for video recognition.
\newblock In {\em Proceedings of the IEEE/CVF international conference on
  computer vision}, pages 6202--6211, 2019.

\bibitem{violet}
Tsu-Jui Fu, Linjie Li, Zhe Gan, Kevin Lin, William~Yang Wang, Lijuan Wang, and
  Zicheng Liu.
\newblock Violet: End-to-end video-language transformers with masked
  visual-token modeling.
\newblock {\em arXiv preprint arXiv:2111.12681}, 2021.

\bibitem{rulstm}
Antonino Furnari and Giovanni~Maria Farinella.
\newblock Rolling-unrolling lstms for action anticipation from first-person
  video.
\newblock {\em IEEE transactions on pattern analysis and machine intelligence},
  43(11):4021--4036, 2020.

\bibitem{aggregation-1}
Adrien Gaidon, Zaid Harchaoui, and Cordelia Schmid.
\newblock Temporal localization of actions with actoms.
\newblock {\em IEEE transactions on pattern analysis and machine intelligence},
  35(11):2782--2795, 2013.

\bibitem{gao2017red}
Jiyang Gao, Zhenheng Yang, and Ram Nevatia.
\newblock Red: Reinforced encoder-decoder networks for action anticipation.
\newblock {\em arXiv preprint arXiv:1707.04818}, 2017.

\bibitem{coot}
Simon Ging, Mohammadreza Zolfaghari, Hamed Pirsiavash, and Thomas Brox.
\newblock Coot: Cooperative hierarchical transformer for video-text
  representation learning.
\newblock {\em Advances in neural information processing systems},
  33:22605--22618, 2020.

\bibitem{avt}
Rohit Girdhar and Kristen Grauman.
\newblock Anticipative video transformer.
\newblock In {\em Proceedings of the IEEE/CVF International Conference on
  Computer Vision}, pages 13505--13515, 2021.

\bibitem{pre-computed-3}
Rohit Girdhar, Deva Ramanan, Abhinav Gupta, Josef Sivic, and Bryan Russell.
\newblock Actionvlad: Learning spatio-temporal aggregation for action
  classification.
\newblock In {\em Proceedings of the IEEE conference on computer vision and
  pattern recognition}, pages 971--980, 2017.

\bibitem{omnivore}
Rohit Girdhar, Mannat Singh, Nikhila Ravi, Laurens van~der Maaten, Armand
  Joulin, and Ishan Misra.
\newblock Omnivore: A single model for many visual modalities.
\newblock In {\em Proceedings of the IEEE/CVF Conference on Computer Vision and
  Pattern Recognition}, pages 16102--16112, 2022.

\bibitem{catastrophic-forgetting-3}
Ian~J Goodfellow, Mehdi Mirza, Da Xiao, Aaron Courville, and Yoshua Bengio.
\newblock An empirical investigation of catastrophic forgetting in
  gradient-based neural networks.
\newblock {\em arXiv preprint arXiv:1312.6211}, 2013.

\bibitem{ego4d}
Kristen Grauman, Andrew Westbury, Eugene Byrne, Zachary Chavis, Antonino
  Furnari, Rohit Girdhar, Jackson Hamburger, Hao Jiang, Miao Liu, Xingyu Liu,
  et~al.
\newblock Ego4d: Around the world in 3,000 hours of egocentric video.
\newblock In {\em Proceedings of the IEEE/CVF Conference on Computer Vision and
  Pattern Recognition}, pages 18995--19012, 2022.

\bibitem{gu-s4}
Albert Gu, Karan Goel, and Christopher R{\'e}.
\newblock Efficiently modeling long sequences with structured state spaces.
\newblock {\em arXiv preprint arXiv:2111.00396}, 2021.

\bibitem{captioning-1}
Xiaowei Hu, Zhe Gan, Jianfeng Wang, Zhengyuan Yang, Zicheng Liu, Yumao Lu, and
  Lijuan Wang.
\newblock Scaling up vision-language pre-training for image captioning.
\newblock In {\em Proceedings of the IEEE/CVF Conference on Computer Vision and
  Pattern Recognition}, pages 17980--17989, 2022.

\bibitem{img-retrieval-5}
Yan Huang, Qi Wu, Chunfeng Song, and Liang Wang.
\newblock Learning semantic concepts and order for image and sentence matching.
\newblock In {\em Proceedings of the IEEE Conference on Computer Vision and
  Pattern Recognition}, pages 6163--6171, 2018.

\bibitem{more-frame-1}
Noureldien Hussein, Efstratios Gavves, and Arnold~WM Smeulders.
\newblock Timeception for complex action recognition.
\newblock In {\em Proceedings of the IEEE/CVF Conference on Computer Vision and
  Pattern Recognition}, pages 254--263, 2019.

\bibitem{gedas-eccv22}
Md~Mohaiminul Islam and Gedas Bertasius.
\newblock Long movie clip classification with state-space video models.
\newblock {\em arXiv preprint arXiv:2204.01692}, 2022.

\bibitem{epic-fusion}
Evangelos Kazakos, Arsha Nagrani, Andrew Zisserman, and Dima Damen.
\newblock Epic-fusion: Audio-visual temporal binding for egocentric action
  recognition.
\newblock In {\em Proceedings of the IEEE/CVF International Conference on
  Computer Vision}, pages 5492--5501, 2019.

\bibitem{catastrophic-forgetting-2}
Ronald Kemker, Marc McClure, Angelina Abitino, Tyler Hayes, and Christopher
  Kanan.
\newblock Measuring catastrophic forgetting in neural networks.
\newblock In {\em Proceedings of the AAAI Conference on Artificial
  Intelligence}, volume~32, 2018.

\bibitem{catastrophic-forgetting-1}
James Kirkpatrick, Razvan Pascanu, Neil Rabinowitz, Joel Veness, Guillaume
  Desjardins, Andrei~A Rusu, Kieran Milan, John Quan, Tiago Ramalho, Agnieszka
  Grabska-Barwinska, et~al.
\newblock Overcoming catastrophic forgetting in neural networks.
\newblock {\em Proceedings of the national academy of sciences},
  114(13):3521--3526, 2017.

\bibitem{scsampler}
Bruno Korbar, Du Tran, and Lorenzo Torresani.
\newblock Scsampler: Sampling salient clips from video for efficient action
  recognition.
\newblock In {\em Proceedings of the IEEE/CVF International Conference on
  Computer Vision}, pages 6232--6242, 2019.

\bibitem{img-retrieval-4}
Kuang-Huei Lee, Xi Chen, Gang Hua, Houdong Hu, and Xiaodong He.
\newblock Stacked cross attention for image-text matching.
\newblock In {\em Proceedings of the European conference on computer vision
  (ECCV)}, pages 201--216, 2018.

\bibitem{vqa-2}
Jie Lei, Tamara~L Berg, and Mohit Bansal.
\newblock Revealing single frame bias for video-and-language learning.
\newblock {\em arXiv preprint arXiv:2206.03428}, 2022.

\bibitem{clipbert}
Jie Lei, Linjie Li, Luowei Zhou, Zhe Gan, Tamara~L Berg, Mohit Bansal, and
  Jingjing Liu.
\newblock Less is more: Clipbert for video-and-language learning via sparse
  sampling.
\newblock In {\em Proceedings of the IEEE/CVF Conference on Computer Vision and
  Pattern Recognition}, pages 7331--7341, 2021.

\bibitem{tvqa+}
Jie Lei, Licheng Yu, Tamara~L Berg, and Mohit Bansal.
\newblock Tvqa+: Spatio-temporal grounding for video question answering.
\newblock {\em arXiv preprint arXiv:1904.11574}, 2019.

\bibitem{uniformer}
Kunchang Li, Yali Wang, Junhao Zhang, Peng Gao, Guanglu Song, Yu Liu, Hongsheng
  Li, and Yu Qiao.
\newblock Uniformer: Unifying convolution and self-attention for visual
  recognition.
\newblock {\em arXiv preprint arXiv:2201.09450}, 2022.

\bibitem{img-retrieval-3}
Kunpeng Li, Yulun Zhang, Kai Li, Yuanyuan Li, and Yun Fu.
\newblock Visual semantic reasoning for image-text matching.
\newblock In {\em Proceedings of the IEEE/CVF international conference on
  computer vision}, pages 4654--4662, 2019.

\bibitem{hero}
Linjie Li, Yen-Chun Chen, Yu Cheng, Zhe Gan, Licheng Yu, and Jingjing Liu.
\newblock Hero: Hierarchical encoder for video+ language omni-representation
  pre-training.
\newblock {\em arXiv preprint arXiv:2005.00200}, 2020.

\bibitem{captioning-3}
Xiujun Li, Xi Yin, Chunyuan Li, Pengchuan Zhang, Xiaowei Hu, Lei Zhang, Lijuan
  Wang, Houdong Hu, Li Dong, Furu Wei, et~al.
\newblock Oscar: Object-semantics aligned pre-training for vision-language
  tasks.
\newblock In {\em European Conference on Computer Vision}, pages 121--137.
  Springer, 2020.

\bibitem{ego-exo}
Yanghao Li, Tushar Nagarajan, Bo Xiong, and Kristen Grauman.
\newblock Ego-exo: Transferring visual representations from third-person to
  first-person videos.
\newblock In {\em Proceedings of the IEEE/CVF Conference on Computer Vision and
  Pattern Recognition}, pages 6943--6953, 2021.

\bibitem{mvitv2}
Yanghao Li, Chao-Yuan Wu, Haoqi Fan, Karttikeya Mangalam, Bo Xiong, Jitendra
  Malik, and Christoph Feichtenhofer.
\newblock Mvitv2: Improved multiscale vision transformers for classification
  and detection.
\newblock In {\em Proceedings of the IEEE/CVF Conference on Computer Vision and
  Pattern Recognition}, pages 4804--4814, 2022.

\bibitem{more-frames-2}
Ji Lin, Chuang Gan, and Song Han.
\newblock Tsm: Temporal shift module for efficient video understanding.
\newblock In {\em Proceedings of the IEEE/CVF International Conference on
  Computer Vision}, pages 7083--7093, 2019.

\bibitem{egovlp}
Kevin~Qinghong Lin, Alex~Jinpeng Wang, Mattia Soldan, Michael Wray, Rui Yan,
  Eric~Zhongcong Xu, Difei Gao, Rongcheng Tu, Wenzhe Zhao, Weijie Kong, et~al.
\newblock Egocentric video-language pretraining.
\newblock In {\em NeurIPS}, 2022.

\bibitem{adamw}
Ilya Loshchilov and Frank Hutter.
\newblock Decoupled weight decay regularization.
\newblock {\em arXiv preprint arXiv:1711.05101}, 2017.

\bibitem{img-grounding-3}
Gen Luo, Yiyi Zhou, Xiaoshuai Sun, Liujuan Cao, Chenglin Wu, Cheng Deng, and
  Rongrong Ji.
\newblock Multi-task collaborative network for joint referring expression
  comprehension and segmentation.
\newblock In {\em Proceedings of the IEEE/CVF Conference on computer vision and
  pattern recognition}, pages 10034--10043, 2020.

\bibitem{univl}
Huaishao Luo, Lei Ji, Botian Shi, Haoyang Huang, Nan Duan, Tianrui Li, Jason
  Li, Taroon Bharti, and Ming Zhou.
\newblock Univl: A unified video and language pre-training model for multimodal
  understanding and generation.
\newblock {\em arXiv preprint arXiv:2002.06353}, 2020.

\bibitem{img-grounding-4}
Junhua Mao, Jonathan Huang, Alexander Toshev, Oana Camburu, Alan~L Yuille, and
  Kevin Murphy.
\newblock Generation and comprehension of unambiguous object descriptions.
\newblock In {\em Proceedings of the IEEE conference on computer vision and
  pattern recognition}, pages 11--20, 2016.

\bibitem{intention}
Esteve~Valls Mascaro, Hyemin Ahn, and Dongheui Lee.
\newblock Intention-conditioned long-term human egocentric action forecasting@
  ego4d challenge 2022.
\newblock {\em arXiv preprint arXiv:2207.12080}, 2022.

\bibitem{mil-nce}
Antoine Miech, Jean-Baptiste Alayrac, Lucas Smaira, Ivan Laptev, Josef Sivic,
  and Andrew Zisserman.
\newblock End-to-end learning of visual representations from uncurated
  instructional videos.
\newblock In {\em Proceedings of the IEEE/CVF Conference on Computer Vision and
  Pattern Recognition}, pages 9879--9889, 2020.

\bibitem{aggregation-2}
Antoine Miech, Ivan Laptev, and Josef Sivic.
\newblock Learnable pooling with context gating for video classification.
\newblock {\em arXiv preprint arXiv:1706.06905}, 2017.

\bibitem{howto100m}
Antoine Miech, Dimitri Zhukov, Jean-Baptiste Alayrac, Makarand Tapaswi, Ivan
  Laptev, and Josef Sivic.
\newblock Howto100m: Learning a text-video embedding by watching hundred
  million narrated video clips.
\newblock In {\em Proceedings of the IEEE/CVF International Conference on
  Computer Vision}, pages 2630--2640, 2019.

\bibitem{naing2020procedure}
Zwe Naing and Ehsan Elhamifar.
\newblock Procedure completion by learning from partial summaries.
\newblock In {\em British Machine Vision Conference}, 2020.

\bibitem{captioning-2}
Van-Quang Nguyen, Masanori Suganuma, and Takayuki Okatani.
\newblock Grit: Faster and better image captioning transformer using dual
  visual features.
\newblock {\em arXiv preprint arXiv:2207.09666}, 2022.

\bibitem{infonce}
Aaron van~den Oord, Yazhe Li, and Oriol Vinyals.
\newblock Representation learning with contrastive predictive coding.
\newblock {\em arXiv preprint arXiv:1807.03748}, 2018.

\bibitem{aggregation-3}
Hamed Pirsiavash and Deva Ramanan.
\newblock Parsing videos of actions with segmental grammars.
\newblock In {\em Proceedings of the IEEE conference on computer vision and
  pattern recognition}, pages 612--619, 2014.

\bibitem{clip}
Alec Radford, Jong~Wook Kim, Chris Hallacy, Aditya Ramesh, Gabriel Goh,
  Sandhini Agarwal, Girish Sastry, Amanda Askell, Pamela Mishkin, Jack Clark,
  Gretchen Krueger, and Ilya Sutskever.
\newblock Learning transferable visual models from natural language
  supervision.
\newblock In Marina Meila and Tong Zhang, editors, {\em Proceedings of the 38th
  International Conference on Machine Learning}, volume 139 of {\em Proceedings
  of Machine Learning Research}, pages 8748--8763. PMLR, 18--24 Jul 2021.

\bibitem{naq}
Santhosh Ramakrishnan, Ziad Al-Halah, and Kristen Grauman.
\newblock Naq: Leveraging narrations as queries to supervise episodic memory.
\newblock In {\em CVPR}, 2023.

\bibitem{brave}
Adria Recasens, Pauline Luc, Jean-Baptiste Alayrac, Luyu Wang, Florian Strub,
  Corentin Tallec, Mateusz Malinowski, Viorica P{\u{a}}tr{\u{a}}ucean, Florent
  Altch{\'e}, Michal Valko, et~al.
\newblock Broaden your views for self-supervised video learning.
\newblock In {\em Proceedings of the IEEE/CVF International Conference on
  Computer Vision}, pages 1255--1265, 2021.

\bibitem{distilbert}
Victor Sanh, Lysandre Debut, Julien Chaumond, and Thomas Wolf.
\newblock Distilbert, a distilled version of bert: smaller, faster, cheaper and
  lighter.
\newblock {\em arXiv preprint arXiv:1910.01108}, 2019.

\bibitem{img-grounding-5}
Hengcan Shi, Hongliang Li, Fanman Meng, and Qingbo Wu.
\newblock Key-word-aware network for referring expression image segmentation.
\newblock In {\em Proceedings of the European Conference on Computer Vision
  (ECCV)}, pages 38--54, 2018.

\bibitem{actor}
Gunnar~A Sigurdsson, Abhinav Gupta, Cordelia Schmid, Ali Farhadi, and Karteek
  Alahari.
\newblock Actor and observer: Joint modeling of first and third-person videos.
\newblock In {\em Proceedings of the IEEE Conference on Computer Vision and
  Pattern Recognition}, pages 7396--7404, 2018.

\bibitem{charades-ego}
Gunnar~A Sigurdsson, Abhinav Gupta, Cordelia Schmid, Ali Farhadi, and Karteek
  Alahari.
\newblock Charades-ego: A large-scale dataset of paired third and first person
  videos.
\newblock {\em arXiv preprint arXiv:1804.09626}, 2018.

\bibitem{coin}
Yansong Tang, Dajun Ding, Yongming Rao, Yu Zheng, Danyang Zhang, Lili Zhao,
  Jiwen Lu, and Jie Zhou.
\newblock Coin: A large-scale dataset for comprehensive instructional video
  analysis.
\newblock In {\em Proceedings of the IEEE/CVF Conference on Computer Vision and
  Pattern Recognition}, pages 1207--1216, 2019.

\bibitem{movieqa}
Makarand Tapaswi, Yukun Zhu, Rainer Stiefelhagen, Antonio Torralba, Raquel
  Urtasun, and Sanja Fidler.
\newblock Movieqa: Understanding stories in movies through question-answering.
\newblock In {\em Proceedings of the IEEE conference on computer vision and
  pattern recognition}, pages 4631--4640, 2016.

\bibitem{aggregation-4}
G{\"u}l Varol, Ivan Laptev, and Cordelia Schmid.
\newblock Long-term temporal convolutions for action recognition.
\newblock {\em IEEE transactions on pattern analysis and machine intelligence},
  40(6):1510--1517, 2017.

\bibitem{Wang_2022_CVPR}
Jue Wang, Gedas Bertasius, Du Tran, and Lorenzo Torresani.
\newblock Long-short temporal contrastive learning of video transformers.
\newblock In {\em Proceedings of the IEEE/CVF Conference on Computer Vision and
  Pattern Recognition (CVPR)}, pages 14010--14020, June 2022.

\bibitem{omnivl}
Junke Wang, Dongdong Chen, Zuxuan Wu, Chong Luo, Luowei Zhou, Yucheng Zhao,
  Yujia Xie, Ce Liu, Yu-Gang Jiang, and Lu Yuan.
\newblock Omnivl: One foundation model for image-language and video-language
  tasks.
\newblock {\em arXiv preprint arXiv:2209.07526}, 2022.

\bibitem{aggregation-5}
Jue Wang and Anoop Cherian.
\newblock Learning discriminative video representations using adversarial
  perturbations.
\newblock In {\em Proceedings of the European Conference on Computer Vision
  (ECCV)}, pages 685--701, 2018.

\bibitem{captioning-5}
Jianfeng Wang, Zhengyuan Yang, Xiaowei Hu, Linjie Li, Kevin Lin, Zhe Gan,
  Zicheng Liu, Ce Liu, and Lijuan Wang.
\newblock Git: A generative image-to-text transformer for vision and language.
\newblock {\em arXiv preprint arXiv:2205.14100}, 2022.

\bibitem{aggregation-6}
Limin Wang, Yuanjun Xiong, Zhe Wang, Yu Qiao, Dahua Lin, Xiaoou Tang, and Luc
  Van~Gool.
\newblock Temporal segment networks: Towards good practices for deep action
  recognition.
\newblock In {\em European conference on computer vision}, pages 20--36.
  Springer, 2016.

\bibitem{wray2019fine}
Michael Wray, Diane Larlus, Gabriela Csurka, and Dima Damen.
\newblock Fine-grained action retrieval through multiple parts-of-speech
  embeddings.
\newblock In {\em Proceedings of the IEEE/CVF International Conference on
  Computer Vision}, pages 450--459, 2019.

\bibitem{ltfb}
Chao-Yuan Wu, Christoph Feichtenhofer, Haoqi Fan, Kaiming He, Philipp
  Krahenbuhl, and Ross Girshick.
\newblock Long-term feature banks for detailed video understanding.
\newblock In {\em Proceedings of the IEEE/CVF Conference on Computer Vision and
  Pattern Recognition}, pages 284--293, 2019.

\bibitem{long-form-video-understanding}
Chao-Yuan Wu and Philipp Krahenbuhl.
\newblock Towards long-form video understanding.
\newblock In {\em Proceedings of the IEEE/CVF Conference on Computer Vision and
  Pattern Recognition}, pages 1884--1894, 2021.

\bibitem{memvit}
Chao-Yuan Wu, Yanghao Li, Karttikeya Mangalam, Haoqi Fan, Bo Xiong, Jitendra
  Malik, and Christoph Feichtenhofer.
\newblock Memvit: Memory-augmented multiscale vision transformer for efficient
  long-term video recognition.
\newblock In {\em Proceedings of the IEEE/CVF Conference on Computer Vision and
  Pattern Recognition}, pages 13587--13597, 2022.

\bibitem{more-frames-3}
Chao-Yuan Wu, Manzil Zaheer, Hexiang Hu, R Manmatha, Alexander~J Smola, and
  Philipp Kr{\"a}henb{\"u}hl.
\newblock Compressed video action recognition.
\newblock In {\em Proceedings of the IEEE conference on computer vision and
  pattern recognition}, pages 6026--6035, 2018.

\bibitem{s3d-g}
Saining Xie, Chen Sun, Jonathan Huang, Zhuowen Tu, and Kevin Murphy.
\newblock Rethinking spatiotemporal feature learning: Speed-accuracy trade-offs
  in video classification.
\newblock In {\em Proceedings of the European conference on computer vision
  (ECCV)}, pages 305--321, 2018.

\bibitem{xu2021vlm}
Hu Xu, Gargi Ghosh, Po-Yao Huang, Prahal Arora, Masoumeh Aminzadeh, Christoph
  Feichtenhofer, Florian Metze, and Luke Zettlemoyer.
\newblock Vlm: Task-agnostic video-language model pre-training for video
  understanding.
\newblock {\em arXiv preprint arXiv:2105.09996}, 2021.

\bibitem{videoclip}
Hu Xu, Gargi Ghosh, Po-Yao Huang, Dmytro Okhonko, Armen Aghajanyan, Florian
  Metze, Luke Zettlemoyer, and Christoph Feichtenhofer.
\newblock Videoclip: Contrastive pre-training for zero-shot video-text
  understanding.
\newblock {\em arXiv preprint arXiv:2109.14084}, 2021.

\bibitem{msr-vtt}
Jun Xu, Tao Mei, Ting Yao, and Yong Rui.
\newblock Msr-vtt: A large video description dataset for bridging video and
  language.
\newblock In {\em Proceedings of the IEEE conference on computer vision and
  pattern recognition}, pages 5288--5296, 2016.

\bibitem{just-ask}
Antoine Yang, Antoine Miech, Josef Sivic, Ivan Laptev, and Cordelia Schmid.
\newblock Just ask: Learning to answer questions from millions of narrated
  videos.
\newblock In {\em Proceedings of the IEEE/CVF International Conference on
  Computer Vision}, pages 1686--1697, 2021.

\bibitem{vqa-1}
Antoine Yang, Antoine Miech, Josef Sivic, Ivan Laptev, and Cordelia Schmid.
\newblock Zero-shot video question answering via frozen bidirectional language
  models.
\newblock {\em arXiv preprint arXiv:2206.08155}, 2022.

\bibitem{taco}
Jianwei Yang, Yonatan Bisk, and Jianfeng Gao.
\newblock Taco: Token-aware cascade contrastive learning for video-text
  alignment.
\newblock In {\em Proceedings of the IEEE/CVF International Conference on
  Computer Vision}, pages 11562--11572, 2021.

\bibitem{pre-computed-4}
Joe Yue-Hei~Ng, Matthew Hausknecht, Sudheendra Vijayanarasimhan, Oriol Vinyals,
  Rajat Monga, and George Toderici.
\newblock Beyond short snippets: Deep networks for video classification.
\newblock In {\em Proceedings of the IEEE conference on computer vision and
  pattern recognition}, pages 4694--4702, 2015.

\bibitem{aggregation-7}
Joe Yue-Hei~Ng, Matthew Hausknecht, Sudheendra Vijayanarasimhan, Oriol Vinyals,
  Rajat Monga, and George Toderici.
\newblock Beyond short snippets: Deep networks for video classification.
\newblock In {\em Proceedings of the IEEE conference on computer vision and
  pattern recognition}, pages 4694--4702, 2015.

\bibitem{merlot}
Rowan Zellers, Ximing Lu, Jack Hessel, Youngjae Yu, Jae~Sung Park, Jize Cao,
  Ali Farhadi, and Yejin Choi.
\newblock Merlot: Multimodal neural script knowledge models.
\newblock {\em Advances in Neural Information Processing Systems},
  34:23634--23651, 2021.

\bibitem{captioning-4}
Yan Zeng, Xinsong Zhang, and Hang Li.
\newblock Multi-grained vision language pre-training: Aligning texts with
  visual concepts.
\newblock {\em arXiv preprint arXiv:2111.08276}, 2021.

\bibitem{img-retrieval-1}
Yan Zeng, Xinsong Zhang, and Hang Li.
\newblock Multi-grained vision language pre-training: Aligning texts with
  visual concepts.
\newblock {\em arXiv preprint arXiv:2111.08276}, 2021.

\bibitem{cmhse}
Bowen Zhang, Hexiang Hu, and Fei Sha.
\newblock Cross-modal and hierarchical modeling of video and text.
\newblock In {\em Proceedings of the european conference on computer vision
  (ECCV)}, pages 374--390, 2018.

\bibitem{actionformer}
Chenlin Zhang, Jianxin Wu, and Yin Li.
\newblock Actionformer: Localizing moments of actions with transformers.
\newblock {\em arXiv preprint arXiv:2202.07925}, 2022.

\bibitem{zhao2022p3iv}
He Zhao, Isma Hadji, Nikita Dvornik, Konstantinos~G Derpanis, Richard~P Wildes,
  and Allan~D Jepson.
\newblock P3iv: Probabilistic procedure planning from instructional videos with
  weak supervision.
\newblock In {\em Proceedings of the IEEE/CVF Conference on Computer Vision and
  Pattern Recognition}, pages 2938--2948, 2022.

\bibitem{vid-retrieval-2}
Shuai Zhao, Linchao Zhu, Xiaohan Wang, and Yi Yang.
\newblock Centerclip: Token clustering for efficient text-video retrieval.
\newblock {\em arXiv preprint arXiv:2205.00823}, 2022.

\bibitem{more-frames-4}
Bolei Zhou, Alex Andonian, Aude Oliva, and Antonio Torralba.
\newblock Temporal relational reasoning in videos.
\newblock In {\em Proceedings of the European conference on computer vision
  (ECCV)}, pages 803--818, 2018.

\bibitem{youcook2}
Luowei Zhou, Chenliang Xu, and Jason~J Corso.
\newblock Towards automatic learning of procedures from web instructional
  videos.
\newblock In {\em AAAI Conference on Artificial Intelligence}, pages
  7590--7598, 2018.

\bibitem{crosstask}
Dimitri Zhukov, Jean-Baptiste Alayrac, Ramazan~Gokberk Cinbis, David Fouhey,
  Ivan Laptev, and Josef Sivic.
\newblock Cross-task weakly supervised learning from instructional videos.
\newblock In {\em Proceedings of the IEEE/CVF Conference on Computer Vision and
  Pattern Recognition}, pages 3537--3545, 2019.

\bibitem{more-frames-5}
Mohammadreza Zolfaghari, Kamaljeet Singh, and Thomas Brox.
\newblock Eco: Efficient convolutional network for online video understanding.
\newblock In {\em Proceedings of the European conference on computer vision
  (ECCV)}, pages 695--712, 2018.

\end{thebibliography}
}

\end{document}